\documentclass[12pt, journal,onecolumn]{IEEEtran}

\usepackage{dsfont}
\usepackage{amsmath} 
\usepackage{amsfonts}
\usepackage{graphicx}
\usepackage{multirow}
\usepackage{subfigure}
\usepackage{comment}
\usepackage{color} 
\usepackage{setspace}

\newcommand{\tr}{\mathsf{tr}}
\newcommand{\dx}[1]{\,\mathsf{d}#1}
\newcommand{\vv}[1]{\ensuremath{{\mathsf{#1}}}} 
\newcommand{\mm}[1]{\ensuremath{{\mathsf{#1}}}} 
\newcommand{\transpose}[1]{\ensuremath{{#1}^{\mathsf{T}}}} 
\newcommand{\set}[1]{\ensuremath{{\mathcal{#1}}}} 
\newcommand{\bb}[1]{\ensuremath{{\mathbb{#1}}}} 
\newcommand{\E}{\ensuremath{{\mathbb{E}}}} 

\doublespace

\title{A Learning Theory in Linear Systems under Compositional Models}

\author{ Se Un Park 
\thanks{Se Un Park is with Schlumberger, Houston, TX 77077, USA. 
(e-mail: spark4@slb.com, seunpark@umich.edu). }
}

\begin{document}


\maketitle


\begin{abstract}
We present a learning theory for the training of a linear system operator having an input compositional variable and propose a Bayesian inversion method for inferring the unknown variable from an output of a noisy linear system. We assume that we have partial or even no knowledge of the operator but have training data of input and ouput. A compositional variable satisfies the constraints that the elements of the variable are all non-negative and sum to unity. We quantified the uncertainty in the trained operator and present the convergence rates of training in explicit forms for several interesting cases under stochastic compositional models. The trained linear operator with the covariance matrix, estimated from the training set of pairs of ground-truth input and noisy output data, is further used in evaluation of posterior uncertainty of the solution. This posterior uncertainty clearly demonstrates uncertainty propagation from noisy training data and addresses possible mismatch between the true operator and the estimated one in the final solution. 
\end{abstract}

{\small {\center \bf Index Terms }
Compositional models, Bayesian inversion, Machine learning, Markov chain Monte Carlo, Semi-blind deconvolution, Uncertainty quantification
}

\section{Introduction}
Compositional models are used universally in interpretation of compositions or relative ratios of discrete sections. A compositional model satisfies the constraints that the model elements are all non-negative and sum to unity. For example, pie charts, populations ratio, financial portfolios, etc. adopt this model. A probability mass function for a discrete random variable can be also considered as a compositional variable.
However, direct inference for compositional variables can be difficult. Instead, transformation methods are often used. For example, softmax functions or similar functions transform the intermediate solution vectors in a multi-dimensional infinite or confined space to compositional variables. The result compositional variables through transformations can be interpreted as probability functions but they are not derived from probability models.
Furthermore, describing the probability space of the inferred compositional solution after transformation is generally difficult, given a likelihood or probability function for an observation. 
This work addresses these two problems. 
We assume that we have partial or even no knowledge of the operator but have training data of input and ouput. We present a training method from ground-truth compositional variables and corresponding noisy outputs from a linear system.  Then, we quantify the uncertainty in the trained operator. For several stochastic compositional models, we explicitly derived the convergence rates. With an estimated covariance matrix, we present a method to evaluate posterior solutions under a Bayesian framework and this addresses a possible mismatch between the true system operator and the estimated one. The posterior density for the compositional variable clearly demonstrates uncertainty propagation coming from the noisy training data.

\section{Problem Statement}

The linear forward model for noisy measurements can be stated as 
\begin{equation}
\label{eq:sys_general}
 \vv{s} = \mm{A}\vv{m} + \vv{n},
\end{equation}
where $\vv{m} \in \mathbb{R}^M$ is a compositional vector of fractions, $\vv{s} \in \mathbb{R}^T$ is an observation vector, $\vv{n}  \in \mathbb{R}^T$ is an additive noise, and $\mm{A}  \in \mathbb{R}^{T\times M}$ is a matrix that describes the generative process. 
Assuming a zero-mean multinormal noise model with covariance matrix $\mm{\Sigma}$, the likelihood function that relates a measurement with the input fractions is given by
\begin{equation}
\label{eq:likelihood}
p(\vv{s}|\vv{m}, \mm{A}) = \frac{1}{\sqrt{(2\pi)^T|\mm{\Sigma}|}}\exp\left[-\frac{1}{2}
\transpose{\left(\vv{s} - \,\mm{A}\vv{m}\right)}{\mm{\Sigma}}^{-1}\left(\vv{s} - \,\mm{A}\vv{m}\right)\right],
\end{equation}
where %
the superscript $\mathsf{T}$ denotes the transpose operator, and $\mm{\Sigma}^{-1}$ denotes the  inverse of the covariance matrix.
The likelihood function is a multivariate probability density function (PDF) that describes the probability of measuring $\vv{s}$ given any particular choice of $\vv{m}$ and $\mm{A}$.  
PDFs are denoted herein by $p(\cdot)$, and conditional PDFs are denoted by $p(\cdot |\cdot)$, where the distributions of the symbols to the left of `$|$' are conditional on the values of the symbols to the right.
The likelihood function is also conditioned on a number of other aspects, including the form of the forward model, the sample set used to 
train the forward model, and the noise model.
These will be omitted from our notation since we fix these conditions.

Following Bayesian inverse theory, Bayes' theorem \cite{Silvia2006} is applied to reverse the sense of conditionality in the likelihood function to yield an estimate of \vv{m} given \vv{s} and \mm{A}:
\begin{equation}
\label{eq9}
p(\vv{m}|\vv{s} , \mm{A}) = \frac{p(\vv{s}|\vv{m}, \mm{A})p(\vv{m})}{p(\vv{s})}.
\end{equation}
$p(\vv{m})$ is called the prior distribution, and describes any knowledge about $\vv{m}$ that is available prior to the measurement of $\vv{s}$.  
In this work, this prior includes constraints on \vv{m} such that that its elements are non-negative and that the fractions sum to unity. 
$p(\vv{m})$ could also describe correlations between the fractions, which can be learned from a training set. 
$p(\vv{s})$ describes the probability of measuring \vv{s} with respect to the forward model when considering all possible values of \vv{m}.
Since $p(\vv{s})$ is constant for a given forward model, likelihood and prior, it can be taken as a constant of proportionality that normalizes $p(\vv{m}|\vv{s}, \mm{A})$ such that it integrates to unity, and thus it is often not computed in Bayesian inversion.  
However, it plays a role in computing the Bayesian evidence, the measure of model fitness that we will investigate in future \cite{Silvia2006}. %

 $p(\vv{m}|\vv{s}, \mm{A})$ is called the posterior distribution. %
It describes all the information on \vv{m} that can be inferred from \vv{s} and \mm{A}, given all of the assumptions on which it is conditioned.
Regarding the uncertainty of \mm{A}, we can use  the joint posterior distribution $p(\vv{m}, \mm{A}|\vv{s})$. Because we focus on the unknown compositional variable and its uncertainty, we derive a marginalized posterior distribution $p(\vv{m} |\vv{s})$ for \vv{m} after marginalizing \mm{A} from the joint distribution. 
We can draw an ensemble from $p(\vv{m} |\vv{s})$ to provide a representation of the uncertainty in the posterior estimate of \vv{m}.

\section{Training}
\subsection{Estimating \mm{A}}

We can train \mm{A} from a training set of $N$ samples.
Each sample is assumed to have $M$ fractions and the sum of elements in each sample is one.
We denote the set $\set{M}= \{\vv{m}_1,\ldots,\vv{m}_N\}$.
Each compositional vector $\vv{m}_i$ is of length $M$. 
The measurements are obtained for each sample and assigned to the set $\set{S}= \{\vv{s}_1,\ldots,\vv{s}_N\}$, which is the training set for \mm{A}.

Since we are now focusing on the estimation of \mm{A}, the likelihood function (\ref{eq:likelihood}) is now made explicitly conditional on \mm{A}.
In order to estimate \mm{A} from the training set, we apply Bayes' theorem to find the posterior on \mm{A} given the measurements:
\begin{align}
p(\mm{A}|\set{S},\set{M}) &= \frac{p(\set{S}|\set{M}, \mm{A}) p(\mm{A}|\set{M})}{p(\set{S|\set{M}})}\\
&\propto p(\set{S}|\set{M}, \mm{A}) p(\mm{A}|\set{M}),
\end{align}
where the proportionality constant, $p(\set{S}|\set{M})$, is dropped because it is independent of \mm{A}.
Here $p(\mm{A}|\set{M})$ is the prior on \mm{A}, and the likelihood function is given by
\begin{equation}
p(\set{S}|\set{M}, \mm{A})=\frac{1}{\sqrt{(2\pi)^N|\mm{\Sigma}|}}\exp\left[-\frac{1}{2}\sum_{i=1}^N{\left(\vv{s}_i - \mm{A}\vv{m}_i\right)\mm{\Sigma}^{-1}\left(\vv{s}_i - \mm{A}\vv{m}_i\right)}\right].
\end{equation}

Assuming a uniform prior for $p(\mm{A}|\set{M})$, the maximum a posteriori (MAP) estimator of \mm{A} is equivalent to a maximum-likelihood estimator (MLE) for \mm{A}, which is given by the value of \mm{A} that minimizes
\begin{equation}
f(\mm{A})=\frac{1}{2}\sum_{i=1}^N{\left(\vv{s}_i - \mm{A}\vv{m}_i\right)}{\mm{\Sigma}}^{-1}\left(\vv{s}_i - \mm{A}\vv{m}_i\right).
\end{equation}
Taking the derivatives of $f$ with respect to the elements of \mm{A} \cite[Eq.~88]{matrixcookbook} and equating to zero yields the following equation for \mm{A}:
\begin{equation}
-\mm{\Sigma}^{-1}\sum_{i=1}^N{(\vv{s}_i - \mm{A}\vv{m}_i)\transpose{\vv{m}_i}}=\mm{0},
\end{equation}
where \mm{0} is the matrix of zeros of the same dimensions as \mm{A}.   
Since $\mm{\Sigma}$ is the true covariance matrix and not an estimate based on a small number of samples, it has full rank and is thus invertible. 
The above equation can thus be simplified to 
\begin{equation}
\mm{A}\sum_{i=1}^N{\vv{m}_i\transpose{\vv{m}_i}}=\sum_{i=1}^N{\vv{s}_i\transpose{\vv{m}_i}}.
\end{equation}
The formal MAP solution for \mm{A} can thus be expressed in matrix form as
\begin{equation}
\label{eq:Aopt}
\tilde{\mm{A}}=\mm{S}\transpose{\mm{M}}(\mm{M}\transpose{\mm{M}})^{-1},
\end{equation}
where $\mm{M}$ is the matrix whose $i$-th column is $\vv{m}_i$, $\mm{S}$ is the matrix whose $i$-th column is $\vv{s}_i$, and it is assumed that the training set is constructed such that the inverse of $\mm{M}\transpose{\mm{M}}$ exists.
If this inverse does not exist, then we require more samples in the training set.

\subsection{Variance Analysis of the MAP Estimator of \mm{A} }
\label{ssec:vrf}

The MAP estimator (\ref{eq:Aopt}), under our assumption of a uniform prior on \mm{A}, converges with increasing $N$ to the true unknown $\mm{A}$ since it is also a maximum likelihood estimator.
We investigate this convergence rate under the assumption that the noise is multinormal with covariance \mm{\Sigma}. 
In other words, we perform an analysis of variance of the MAP estimator. 
A natural extension of this discussion is to design an experiment of fabricating \mm{M} such that the noise in $\tilde{\mm{A}}$ is optimally reduced.

Since we assume that the noise in each column of \mm{S} is independent and follows a multivariate normal distribution with mean $\vv{0}$ and variance $\mm{\Sigma}$, i.e., $\mathcal{N}(\vv{0}, \mm{\Sigma})$,
 the $i$-th column of \mm{S} is distributed as ${\vv{s}}_i\sim\mathcal{N}(\mm{A} \vv{m}_i,\mm{\Sigma})$.
Denoting the $i$-th column of \mm{B} by $\vv{b}_i$, the distribution of $i$-th column of $\tilde{\mm{A}}$ is given by
\begin{equation}
\label{eq4}
\tilde{\vv{a}}_i\sim\mathcal{N}(\vv{a}_{i},\|\vv{b}_i\|^2\mm{\Sigma}),
\end{equation}
where $\vv{a}_{i}$ is the $i$-th column of $\mm{A}$.
Thus, the estimator $\tilde{\vv{a}}_i$ is unbiased and expectedly centered at the true unknown value  $\vv{a}_{i}$ and 
the variance scaling factor $r_i$ for $\tilde{\vv{a}}_i$ is given by $r_i=\|\vv{b}_i\|^2$, where $\|\cdot\|$ is the Euclidean norm.

Further analyzing this variance scaling factor,
\begin{align}
\label{eq:r_derivation1}
r_i &= \| \vv{b}_i \|^2 = \| \transpose{\mm{M}}(\mm{M}\transpose{\mm{M}})^{-1} \vv{e}_i \|^2 \\
 &= \transpose{\vv{e}}_i (\mm{M}\transpose{\mm{M}})^{-1}  \mm{M}\transpose{\mm{M}} (\mm{M}\transpose{\mm{M}})^{-1} \vv{e}_i \\
\label{eq:r_derivation11}
&= \left[(\mm{M}\transpose{\mm{M}})^{-1} \right]_{i,i}\\
&= \left[ \frac{1}{N} ( \frac{1}{N} \sum_{j=1}^N \vv{m}_j\vv{m}_j^{\mathsf{T}})^{-1} \right]_{i,i}, 
\end{align}
where $\vv{e}_i$ is the $i$-th column of the $M\times M$ identity matrix $\mm{I}_M$, and $[ \cdot]_{i,i}$ is the $i$-th diagonal entry of the enclosed matrix. 
When $N$ is large, by the law of large numbers,
\begin{equation}
\label{eq:r_derivation2}
\left[ \frac{1}{N} \left( \frac{1}{N} \sum_{j=1}^N{\vv{m}_j\vv{m}_j^{\mathsf{T}}}\right)^{-1} \right]_{i,i} \, \rightarrow \,
 \frac{1}{N} \left[ \bb{E} [\vv{m}\transpose{\vv{m}}] \right]_{i,i}^{-1} , 
\end{equation}
where $\bb{E}[\cdot]$ is an expectation of the enclosed matrix. The expectation is defined as
\begin{equation}
\label{eq:expectation}
 \bb{E}[\vv{m}\transpose{\vv{m}}] = \int_{\Omega}{\vv{m}\transpose{\vv{m}}p(\vv{m})\dx{\vv{m}}} , 
\end{equation}
where ${\Omega}$ is the domain for $\vv{m}$, and $p(\vv{m})$ is a probability density function describing the compositions in the training set.
Therefore, assuming that $N$ is large, we have the following value for $r_i$: 
\begin{equation}
\label{eq:r_simple}
r_i = \frac{1}{N} \left[ \bb{E} [\vv{m}\transpose{\vv{m}}] \right]_{i,i}^{-1} .
\end{equation}

The variance scaling factor $r$ can be interpreted as the multiplicative `error'  factor in the estimation of  \mm{A} through training; this is because the true value, which is a mean parameter $\mm{A}$, is unknown in practice, and $ \E[(\tilde{\vv{a}}_i-\vv{a}_{i})\transpose{(\tilde{\vv{a}}_i-\vv{a}_{i})} ] = r_i \mm{\Sigma} $. Thus, the uncertainty in $\tilde{\vv{a}}_i$ is related to the uncertainty in \mm{S} through the scaling factor $r_i$.

In the following two sections we explore the variance scaling factors that result from different choices of compositional mixtures in the training set. 
We first consider the case of repeated identical compositional mixtures, and use (\ref{eq:r_derivation11}) to compute $r_i$.
We next consider the case where the compositional vectors are stochastic, and use (\ref{eq:r_simple}) to compute $r_i$. 
In the stochastic case, the probability density function $p(\vv{m})$ in (\ref{eq:expectation}) describes the distribution used in sampling the compositional variables.

\subsubsection{Variance scaling factor for repeated mixtures}
\label{sssec:vrf_cases}

Assume that there are $C$ unique compositions and that each composition in the training set is repeated $K$ times. 
Putting each of these mixtures into a matrix partition $\mm{M}_0 \in \bb{R}^{M \times C}$, the \mm{M} for the full training set is defined by 
$\mm{M} = [\mm{M}_0, \ldots, \mm{M}_0 ] \in \bb{R}^{M \times N}$, with $ N = C K$.
Then (\ref{eq:r_derivation11}) becomes 
\begin{equation}
\label{eq:r_rep}
r_i = \frac{1}{K}  [(\mm{M}_0\transpose{\mm{M}_0})^{-1} ]_{i,i},
\end{equation}
where the compositions are chosen such that $\mm{M}_0\transpose{\mm{M}_0}$ is invertible.
Thus, the noise variance of the estimated \mm{A} reduces as $1/K$ with increasing repetitions of the compositions. 
Note that this convergence rate for repeated compositions is identical to that when `pure' samples are used, \emph{i.e.,} when $\mm{M}_0$ is the identity matrix, $r_i=1/K$.

The variance scaling factor is given by the diagonal values of $(\mm{M}_0\transpose{\mm{M}_0})^{-1}$. 
As noted above, for pure samples this scaling constant is unity.
Next consider binary mixtures of the form
\begin{equation}
\label{eq:M0-binary}
\mm{M}_0=\left[
\begin{array}{ccccc}
1/2 & 0 & 0 & \ldots & 0\\
1/2 & 1/2 & 0 & \ldots & 0 \\
0   & 1/2 & 1/2 & 0 & \vdots \\
\vdots &  & \ddots & 1/2 & 0 \\
0 & 0 & 0 & 1/2 & 1
\end{array}
\right],
\end{equation}
where adjacent pairs of the $M$ endmembers are mixed in equal mass portions, and the $M$-th endmember is provided in pure form in order to make $\mm{M}_0$ invertible.
In this case the variance scaling factor has the form 
$r_i=\left[4(M-i)+1\right]/K$.
$r_1$, the maximum value of $r_i$, varies with $M$ as $(4M-3)/K = (4M-3)M/N $. Also note that the minimum factor is $r_M = 1/K = M/N$. 
For example, with ten endmembers the maximum value is $r_1=37/K$, meaning that the variance is 37 times larger than for a sample of pure compositions.

\subsubsection{Variance scaling factor for stochastic mixtures}

For a stochastic mixture, each column of $\mm{M}$ is a realization of a random vector $\vv{m} = \transpose{[U_1, U_2, \ldots, U_M]}$  subject to a probability density function $p(\vv{m})$ that must honor the non negativity and summation constraints on \vv{m}. 
In practice, the variance scaling factor, $r_i$, could be evaluated by means of Monte Carlo methods for any feasible distribution by simply sampling from that distribution and evaluating the expectation of the function of those samples given in (\ref{eq:r_simple}).
In the following analysis we consider an important special case in which 
\begin{align}
\label{eq:E_assumption3}
\bb{E} [U_j^2] = \bb{E} [U_k^2] \mbox{ and }
\bb{E} [U_j U_k] = \bb{E} [U_l U_m], \mbox{ for } j \neq k , l \neq m.
\end{align}
Under these conditions we have derived simple formulas for $r_i$ for several distributions.
Following the formula for $r_i$ in (\ref{eq:r_simple}) and using the assumption in (\ref{eq:E_assumption3}), the expectation is given by
\begin{equation}
\label{eq:EmmT}
\bb{E}  [\vv{m}\transpose{\vv{m}}] = \alpha \mm{I} + \beta \vv{1}\transpose{\vv{1}},
\end{equation}
where $\alpha = \bb{E} [U_1^2] - \bb{E} [U_1 U_2]$ and $\beta = \bb{E} [U_1 U_2]$.
By using the Sherman-Morrison matrix inversion formula \cite{matrixcookbook},
\begin{equation}
\label{eq:EmmT_inv}
\left( \bb{E}  [\vv{m}\transpose{\vv{m}}] \right)^{-1} = \frac{1}{\alpha} \left[ \mm{I} - \mm{1}\transpose{\mm{1}} \frac{\beta}{\alpha+\beta M } \right] .
\end{equation}
Note that this result holds for a general random vector $\vv{m}$ that satisfies (\ref{eq:E_assumption3}) without positivity or summation constraints.
Therefore, considering the $i$-th diagonal entry of the inverse matrix, 
\begin{align}
\label{eq:vrf}
r_i = r = \frac{1}{N \alpha} \frac{\alpha + (M-1)\beta}{\alpha + M\beta} .
\end{align}
Due to the symmetry assumptions in (\ref{eq:EmmT}), all $r_i$ are identical.
Note that $\alpha$ and $\beta$ are functions of $M$.

In Appendix~\ref{appen:distSimplex}, we derive formulas for $r_i$ for several distributions that satisfy the non negativity and summation constraints on \vv{m}:
\begin{itemize}
\item {\bf Multinomial distribution:} Each sample from this distribution uniformly chooses one pure endmember, with replacement, from the set of $M$ endmembers \\
\item {\bf Double-multinomial distribution with or without replacement:} Each sample from this double-multinomial distribution uniformly chooses two pure endmembers, with or without replacement, from the set of $M$ endmembers\\
\item {\bf Uniform distribution over $M$-dimensional simplex:} Each sample from this distribution uniformly chooses $M$ fractions, subject to the non negativity and summation conditions
\end{itemize}
Formulas for the $r_i$ in these cases are presented in Table~\ref{tab:r} along with the formulas for deterministic mixtures derived above.

\begin{table}
\begin{center}
\begin{tabular}{|l|c|}
\hline
{\bf Mixture} & $\mathbf{r}$ \\ \hline\hline
Repeated Pure Endmembers& $\frac{M}{N}$\rule{0pt}{3ex} \\
Repeated Binary Mixtures & $\frac{M}{N} (4M-3)$\rule{0pt}{3ex}  \\
Multinomial Distribution & $\frac{M}{N}$\rule{0pt}{3ex}  \\
Double Multinomial with Replacement &  $\frac{1}{N} (2 M - 1)$\rule{0pt}{3ex}  \\
Double Multinomial without Replacement &  $\frac{1}{N} \left(2 M + \frac{M}{M-2}\right) $\rule{0pt}{3ex}  \\
Uniform Distribution & $\frac{M^2}{N}$ \\
\hline
\end{tabular}
\end{center}
\caption[]{Variance scaling factors for several types of mixtures. The first two ``repetitive'' cases consider deterministic structures of sample set. The remaining four cases consider expectation from the stochastic compositions according to the given distribution. The worst case is presented for the repetitive binary mixture. For all other cases, the structures of $\mm{M}$ are symmetric so that $r_i$ are all identical. }
\label{tab:r}
\end{table}

\subsubsection{Discussion on the impact of mixture selection in variance reduction }

We present the several variance scaling factors for several mixture selection/preparation strategies in Table~\ref{tab:r}. 
The variance of the linear operator $\tilde{\mm{A}}$ always decreases with the increasing number of training samples $N$ as $1/N$. %
However, it grows with the increasing number of endmembers $M$ with rates that vary with the choice of stochastic mixture.
The worst cases   increase quadratically with $M$; these are the deterministic binary mixtures and the uniform stochastic mixtures.
The best cases increase linearly with $M$; these are the repeated pure endmember sets and the multinomial and double-multinomial mixtures. %

\begin{figure}
  \centering
\begin{tabular}{rl}
  \includegraphics[width=3.6in]{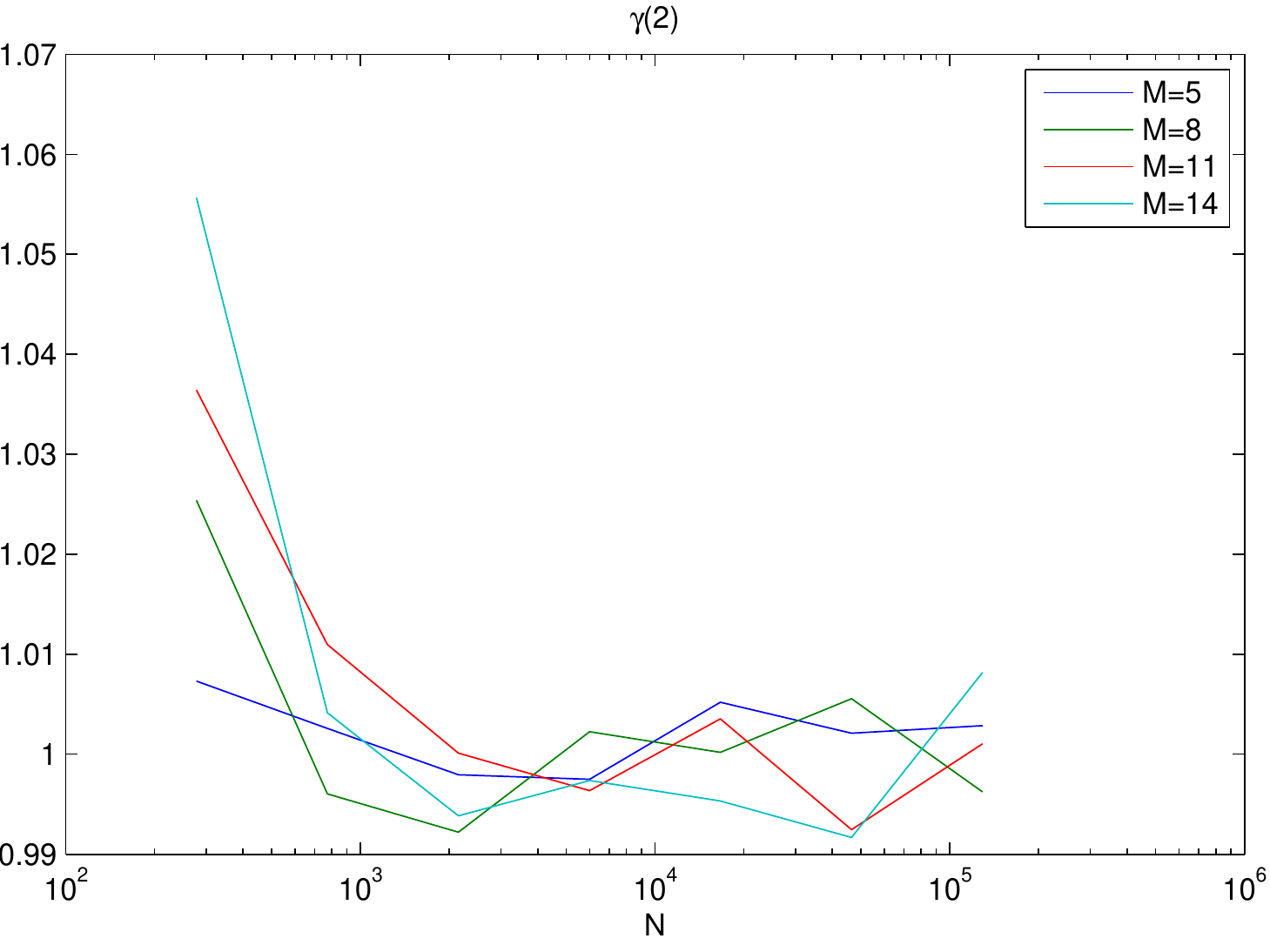}
\end{tabular}
\caption{\label{fig:multiCurves} Under the multinomial distribution on $M$ simplex (see Appendix \ref{appen:distSimplex}), 
 $\gamma(2)$ curves for several $M$ values. Note the convergence after $N \approx 800$.
}
\end{figure}

We verified all the obtained variance scaling factors presented in Table~\ref{tab:r} with numerical simulations. 
For the $j$-th endmember and $t$-th observation index, we define a quantity $\gamma_j(t)$ as the ratio of the estimated variance and the theoretical variance:
\begin{equation}
\gamma_j(t) = \frac{ [Var(\tilde{\vv{a}}_j)]_t }{ r_j \, [\mm{\Sigma}]_t } ,
\end{equation}
where $Var(\cdot)$ is the unbiased covariance estimator that is numerically computed. We applied two noise levels $\sigma_1 = 0.05$ for indices $t \in I_1 = \{1, \ldots , 60\}$ and $\sigma_2 = 0.5$ for observation indices $t \in I_2 = \{61, \ldots , 100\}$, we compute two versions of $\gamma_j$  defined as follows:
\begin{align}
\gamma_j(I_k) =\frac{1}{r_j \, [\mm\Sigma]_t} \left( \frac{1}{|I_k|}  \sum_{t' \in I_k} [Var(\tilde{\vv{a}}_j)]_{t'}  \right)     \mbox{ for } k = 1, 2 ,
\end{align}
where $[\mm\Sigma]_t  = \sigma^2_1 \bb{I}(t \in I_1) + \sigma^2_2 \bb{I}(t \in I_2)$, $\bb{I}$ is an indicator function, and $|I|$ is a cardinality of the set $I$. 
For the multinomial case, we present these $\gamma$ curves in Fig.~\ref{fig:multiCurves} as an illustration of how large $N$ should be, possibly depending on $M$, in order to satisfy the law of large numbers that was used in the derivation of (\ref{eq:r_simple}). 
The quantity $\gamma\approx 1$ indicates that $N$ is large enough for (\ref{eq:r_simple}) to be valid. 
From this figure we see that one-percent accuracy is achieved in this example case when $N > 800$.

\subsubsection{Intuition to best configure training sets}

So far, 
we have investigated the behavior of the estimation variance in several cases. 
Interestingly, a theoretical approximation also gives us insight on the behavior of estimation variance. Rather than focusing on an individual variance scaling factors $r_i$,  consider the sum of these factors, from which we gain insight on how to find $\mm{M}$ that would minimize the estimation variance. The sum of the scaling factors is 
\begin{align}
\tilde r &= \sum_{i=1}^M r_i \nonumber\\
\label{eq:tr1}
&= \tr(\left[(\mm{M}\transpose{\mm{M}})^{-1} \right] \\
\label{eq:tr2}
&= \tr\left[ (\sum_{i=1}^N \vv{m}\transpose{\vv{m}})^{-1} \right],
\end{align}
where $\tr(\cdot)$ is the trace operator, defined as the sum of the diagonal elements of the enclosed matrix. Minimizing (\ref{eq:tr1}) with respect to $\mm{M}$ is not trivial, but optimization of the following quantity is easy and can provide an intuition on the optimization of the scaling factor with the extreme \vv{m}:
\begin{align}
\label{eq:tr4}
\tr(\sum_{i=1}^N \vv{m}\transpose{\vv{m}}) = \sum_{i=1}^N \| \vv{m} \|^2 ,
\end{align}
where we used properties of the trace operator \cite{matrixcookbook}. 
The right-hand side of (\ref{eq:tr4}) indicates that for $\vv{m}$ in the simplex (see Appendix \ref{appen:distSimplex}), the maximum is obtained when $\vv{m} = \vv{e}_j$ for any $j$, corresponding to pure samples, and the minimum is obtained when $\vv{m} = \frac{1}{M} \vv{1}_M$.
The matrix inside $\tr$ in (\ref{eq:tr2}) is the inverse matrix of that inside $\tr$ in (\ref{eq:tr4}), so we might guess that this maximization argument could apply to minimization of the $\tilde r $, and vice versa. Indeed, the collection of maximizers $\vv{e}_j$ of (\ref{eq:tr4}), thus similar to multinomial case, is the minimizer of $\tilde r $ having $M/N$ rate. The minimizer $\frac{1}{M} \vv{1}_M$ of (\ref{eq:tr4}) is the expectation of a random vector $\vv{m}$ when its distribution is uniform, and this uniform distribution is the worst case (maximizer) having quadratic rate $M^2/N$.

\subsection{Estimating the Noise Covariance and its Inverse}
\label{ssec:noiseCov}

The noise covariance matrix plays the role of a weighting factor in the likelihood function.  
Any observational indices that are less conformant to the linear model will be poorly predicted by the linear model and can be downweighted by assuming larger noise variance to these indices.  
In this context, the covariance matrix contributes significantly to an improved estimation of \vv{m}.

Given our training set, the maximum-likelihood estimate of the noise covariance matrix is found from the residuals between the predicted and measured ouput measurements:
\begin{equation}
\label{eq8}
\tilde{\mm{\Sigma}} = \frac{1}{N} \sum_{i=1}^N{\vv{r}_i\transpose{\vv{r}_i}},
\end{equation}
where the residuals, $\vv{r}_i$ are defined by $\vv{r}_i = \vv{s}_i - \tilde{\mm{A}}\vv{m}_i$.  
The tilde over \mm{A} serves as a reminder that $\tilde{\mm{A}}$ is an estimate of \mm{A}.
Through (\ref{eq8}), the uncertainty in $\tilde{\mm{A}}$ is consolidated into $\tilde{\mm{\Sigma}}$. 
Note that the sample covariance matrix has a factor $N-1$ instead of $N$ and we call it $\tilde{\mm{\Sigma}}^{sample} = \left( \tilde\sigma^{sample}_{ij} \right) $.
The correlation coefficient between the observation index $i$ and $j$ is defined as \begin{equation}
\label{eq:corr}
\rho_{i,j} = \tilde{\mm{\Sigma}}_{i,j}/\sqrt{\tilde{\mm{\Sigma}}_{i,i} \tilde{\mm{\Sigma}}_{j,j} },
\end{equation}
where $\tilde{\mm{\Sigma}}_{i,j}$ indicates the value at $i$-th row and $j$-th column of $\tilde{\mm{\Sigma}}$. 

In practice, estimating the noise covariance can be posed as a well-known problem of high dimensional covariance matrix estimation;   
$N$ is often much less than the dimension of \vv{s}, making $\tilde{\mm{\Sigma}}$ rank deficient.  
This rank deficiency of $\tilde{\mm{\Sigma}}$ impacts our estimation of \vv{m} because the likelihood function (\ref{eq:likelihood}) is expressed in terms of the inverse of \mm{\Sigma}. 
Recognizing that the off-diagonal elements of \mm{\Sigma} are poorly estimated when the training set contains a small number of samples, we propose that only a certain portion of the entries close to diagonal entries, or simply just the diagonal entries, of $\tilde{\mm{\Sigma}}$ be estimated.
Thus, the impact of these spurious off-diagonal entries can be offset by accepting only the diagonal elements.
A further advantage of this diagonalization is that the inverse of $\tilde{\mm{\Sigma}}$ is now well defined, which is used in estimation of $\vv{m}$. However, in general this diagonalized version of $\tilde{\mm{\Sigma}}$ may overestimate the uncertainty in \vv{s} because the correlations are removed by marginalizing them away.

More general approaches in high dimensional covariance matrix estimation than the above diagonalization approach include banding \cite{Bickel2008}, tapering \cite{Cai2010}, and block banding \cite{Cai2012} methods. Among these, tapering and block banding require additional knowledge such as the tapering parameter and block structure. The banding method requires only the width of the diagonal band. It assumes a simple model describing a noise structure, because one can expect a nonzero correlation between close observation indices and a zero correlation between far observation indices. %

The banding estimator is defined as 
\begin{equation}
\label{eq:bandingDef}
\hat{\mm{\Sigma}} = \left( \tilde\sigma^{sample}_{ij} \bb{I}( |i-j| \leq K ) \right).
\end{equation}
Under the several conditions \cite{Bickel2008}, the convergence rate is 
\begin{equation}
\label{eq:bandingRate}
\bb{E}  \| \hat{\mm{\Sigma}} - {\mm{\Sigma}} \|^2 \leq C \left( \frac{\log T}{N}\right)^\frac{\alpha}{\alpha+1} , 
\end{equation}
where $C$ is a positive constant, $T$ is the dimension of the output, e.g. ${\mm{\Sigma}} = (\sigma_{ij})$ is a $T \times T$ matrix, and $\alpha$ is the decaying factor of off-diagonal entries such that
\begin{equation}
\label{eq:bandingAlpha}
| \sigma_{ij}  | \leq M |i-j|^{-(\alpha+1)}
\end{equation}
for some $M$ and $\forall i \neq j$.

\section{Inversion}

\label{ssec:InverseProblem}

Once \mm{A} and $\mm{\Sigma}$ have been estimated from the training set of samples, a new output measurement, \vv{s}, can be inverted to create a Bayesian estimate of the associated \vv{m} using (\ref{eq9}). 
In the following, we drop the tildes from \mm{A} and \mm{\Sigma} for notational simplicity and denote the mean of \mm{A} by $\mm{A}_0$.

In the composition model, the prior distribution $p(\vv{m})$ must enforce $\vv{0} \leq\vv{m}\leq \vv{1}$ on the elements of \vv{m} and $\|\vv{m}\|_1 = 1$, \emph{i.e.,} the elements of \vv{m} sum to unity.  
Within the support imposed by these constraints, we assume a uniform distribution for $\vv{m}$, thus noninformative, as its prior distribution $p(\vv{m})$.

\subsection{Solution with $\mm{A}$ fixed}
\label{sssec:solAfixed}
With \mm{A} fixed, 
combining the likelihood function of (\ref{eq:likelihood}) with a prior $p(\vv{m})$ yields a posterior estimate of \vv{m}:
\begin{equation}
\label{eq:post2}
p(\vv{m}|\vv{s}) \propto p(\vv{m})\exp\left[-\frac{1}{2}
\transpose{\left(\vv{s} - \mm{A}\vv{m}\right)}{\mm{\Sigma}}^{-1}\left(\vv{s} - \mm{A}\vv{m}\right)\right],
\end{equation}
with $p(\vv{m}) \propto \bb{I} (\vv{m} \in S_M) $ for an $M$-dimensional simplex $S_M$ (see Appendix  \ref{appen:distSimplex}).
The constant of proportionality is found by normalizing $p(\vv{m}|\vv{s})$ such that it integrates to unity.

The MAP solution of (\ref{eq:post2}) is then given by the value of $\vv{m}$ that satisfies
\begin{equation}
\label{eq:mapeq}
\transpose{\mm{A}}\mm{\Sigma}^{-1}\mm{A}\vv{m}
= \transpose{\mm{A}}\mm{\Sigma}^{-1}\vv{s},
\end{equation}
subject to the non-negativity and summation constraints.
This has the form of a standard linear programming problem that can be solved via standard numerical libraries.
One such numerical solver is the Matlab \cite{MATLAB2012} software toolbox fmincon for solving the linear optimization with linear constraints. 
The solution to (\ref{eq:mapeq}) is guaranteed to be unique when $\transpose{\mm{A}}\mm{\Sigma}^{-1}\mm{A}$ has full rank, as was the case in all of our experiments.

However, even when this MAP solution is unique, there may be strong correlations between elements of \vv{m}.
This indicates that some elements are either fundamentally unresolvable or only weakly resolvable by the linear system.
Note that the summation constraint in the prior imposes a correlation between endmembers even when no correlation is imposed by the likelihood function.
For example, when the composition vector contains only two endmembers, $\vv{m}_1$ and $\vv{m}_2$, %
as $\vv{m}_1$ increases, %
$\vv{m}_2$ must decrease commensurately to preserve the summation constraint.
It is critical to capture this correlation in the analysis to avoid mischaracterization of comnpositional variables.

The uncertainty of \vv{m} cannot be simply described,  as for Gaussian distributions, by the inverse of the Hermitian matrix $\transpose{\mm{A}}\mm{\Sigma}^{-1}\mm{A}$ %
because the constraints impose the boundary of feasible vectors on the posterior.  %
To describe the posterior uncertainty, we use the drawn samples from the posterior distribution 
via Monte Carlo sampling methods. 
One can perform sampling based on a truncated multinormal sampler handling (in-)equality constraints.

\subsection{Solution with stochastic $\mm{A}$}
\label{sssec:solAstoch}

In this section, instead of using a point estimate of \mm{A} ignoring its uncertainty, we assume \mm{A} has stochastic perturbations centered on the MLE of \mm{A}. By following the Bayesian rule, we have the joint posterior distribution for \vv{m} and \mm{A} but \mm{A} will be integrated out so that the final solution for \vv{m} has the uncertainty terms coming from the training of \mm{A}. We present practical estimation methods at the end of this section using Markov chain Monte Carlo techinques.

Assuming the Gaussian noise and independent sample acquisition, a noise in a sample is statistically independent from the noise in another sample, i.e., 
the noise $\vv{n}_k$ in $\vv{s}_k$ from the training set is independent of $\vv{n}_l$ in $\vv{s}_l$. 
The prior distribution of, or the MAP estimator for, $\mm{A}$ can be represented\footnote{Strictly speaking, due to our derivation through the MAP estimator, \mm{A} is a function of $\vv{n}_i$ for $i=1,\ldots,N$ or \mm{S}, thus it is a random matrix.}  as 
\begin{equation}
\label{eq:priorA}
p(\mm{A} ; \mm{M}, \mm{S}  ) \propto  \exp \left[-\frac{1}{2}\tr( \transpose{(\mm{A}-\mm{A}_0)}\mm{C} (\mm{A} -\mm{A}_0) \mm{F}) \right] ,
\end{equation}
where $\mm{A}_0 $ is the mean of $\mm{A}$, $\mm{C}=\mm{\Sigma}^{-1}$ and $\mm{F}$ is a diagonal matrix having $r_j^{-1}$ for $j=1 , \ldots, M$.
Henceforth, we omit the set of the training inputs and measurements, $\{ \mm{M} , \mm{S} \}$, to simplify notations of $p(\cdot)$. 
Since we do not know $\mm{A}_0$, in practice we set  $\mm{A}_0 := \mm{SB} $, which is the MLE solution for $\mm{A}$ with given $\mm{M}$ and $\mm{S}$ and converges to the true mean of \mm{A} as $N$ increases.

The full posterior distribution is then 
\begin{equation}
\label{eq:postAm}
p(\vv{m},\mm{A} | \vv{s} ) = p(\vv{m} | \mm{A},\vv{s})  p(\mm{A} ) .
\end{equation}
And we integrate it with respect to \mm{A} to incorporate the uncertainty coming from  the training of \mm{A}.
\begin{equation}
\label{eq:marignal_m}
p(\vv{m}| \vv{s} ) = \int p(\vv{m},\mm{A} | \vv{s} ) d \mm{A}  = \int p(\vv{m} | \mm{A},\vv{s})  p(\mm{A} ) d\mm{A} . 
\end{equation}
We note that 
\begin{align*}
& \int \exp\left[-\frac{1}{2}
\transpose{\left(\vv{s} - \mm{A}\vv{m}\right)}{\mm{\Sigma}}^{-1}\left(\vv{s} - \mm{A}\vv{m}\right)\right] \exp \left[ -\frac{1}{2}\tr( \transpose{(\mm{A}-\mm{A}_0)}\mm{C} (\mm{A} -\mm{A}_0) \mm{F}) \right] d\mm{A} \\
& = \frac{d}{b^{T/2}}  \exp \left[ -\frac{1}{2} \transpose{(\vv{s}-\mm{A}_0 \vv{m})} \frac{\mm{C}}{b}  (\vv{s}-\mm{A}_0 \vv{m}) \right] ,
\end{align*}
where $b := b(\vv{m}) = 1 + \sum_{j=1}^M m_j^2 r_j ,  d = \left( \prod_{j=1}^M \det(  2\pi{r_j} \mm{\Sigma} ) \right)^{\frac{1}{2}}$. 
Therefore, the posterior marginal distribution for $\vv{m}$
\begin{equation}
\label{eq:marignal_m}
p(\vv{m}| \vv{s} ) \propto \frac{1}{b(\vv{m})^{T/2}}  \exp \left[ -\frac{1}{2} \transpose{(\vv{s}-\mm{A}_0 \vv{m})} (b(\vv{m})\mm{\Sigma})^{-1}  (\vv{s}-\mm{A}_0 \vv{m}) \right] \cdot \bb{I} (\vv{m} \in S_M) .
\end{equation}
This is not generally convex so is difficult to optimize. However, when $r < 1$ and $\vv{m} \in S_M$  it locally looks like a convex function of $\vv{m}$. Also, because $m_j \in [0,1]$ we can obtain the tight maximum bound for $r$ such that (see Appendix \ref{appen:bMax})
\begin{equation}
\label{eq:bIneq}
b  \leq 1 + \max(r_j) =: b_{max} .
\end{equation}
 
When $m_j$ and $r_j$ are small, $b$ is close to one, so the posterior marginal distribution $p(\vv{m}| \vv{s})$ behaves similarly to the case in the previous section, where the uncertainty of $\mm{A}$ is not considered, e.g. the distribution for $\vv{s}$ is a linearly-confined Gaussian distribution with mean $\mm{A}_0\vv{m}$ and variance $\mm{\Sigma}$ and the unbounded version variance for \vv{m} is  $(\transpose{\mm{A}_0 }\mm{\Sigma}^{-1} \mm{A}_0 )^{-1}$ but \vv{m} $\in S_M$. %

The added uncertainty coming from $\mm{A}$ is the multiplicative factor $b$ and the worst/upper bound is $b_{max}$. Note that, $b > 1 $ always since $r>0$ and $m_j>0$. Thus, the uncertainty coming from $\mm{A}$ always increases the uncertainty in estimating $\vv{m}$.

The optimization of $p(\vv{m}| \vv{s})$ can be done by using software packages, Markov chain Monte Carlo (MCMC) methods \cite{Park2012, Robert2004}, or variational Bayes approaches \cite{Park2014}.
In this paper, we propose a sampling technique in the MCMC framework to quantify uncertainties of $\vv{m}$.  
Because Gibbs' sampling is difficult to apply in this complex distribution (\ref{eq:marignal_m}), 
our sampling strategy uses the Metropolis-Hastings algorithm \cite{Robert2004} with proposal distribution for \vv{m}  having %
variance $\mm{\Sigma}_m  := (\transpose{\mm{A}_0 }\mm{\Sigma}^{-1} \mm{A}_0 )^{-1}$ on a simplex domain. 
The proposal distribution $q$ to draw a new sample $\vv{m}'$ given the previous sample $\vv{m}$ is symmetric as follows
\begin{align*}
\label{eq:proposalQ}
q( \vv{m} \rightarrow \vv{m}') &\propto \mathcal{N} ( \vv{m}-\vv{m}' ; \vv{0}, (\transpose{\mm{A}_0 }\mm{\Sigma}^{-1} \mm{A}_0 )^{-1} ) \cdot \bb{I} (\vv{m}' \in S_M)  \\ & = \frac{1}{Z(\vv{m})} \exp \left[ -\frac{1}{2} \left( \transpose{(\vv{m}-\vv{m}')} (\transpose{\mm{A}_0 }\mm{\Sigma}^{-1} \mm{A}_0 )  (\vv{m}-\vv{m}')  \right) \right]\cdot \bb{I} (\vv{m}' \in S_M),
\end{align*}
where $\mathcal{N}$ is a Gaussian distribution and $Z(\vv{m})$ is a normalization constant and function of $\vv{m}$. %
The acceptance rate $A( \vv{m} \rightarrow \vv{m}')$ is then defined as follows
\begin{equation}
\label{eq:Acc1}
A( \vv{m} \rightarrow \vv{m}') = \min \left(1, acc( \vv{m} \rightarrow \vv{m}') \right) ,
\end{equation}
where 
\begin{align*}
\label{eq:Acc2}
acc( \vv{m} \rightarrow \vv{m}') &= \frac{ p(\vv{m}'|\vv{s}) q( \vv{m}' \rightarrow \vv{m})}{ p(\vv{m}|\vv{s}) q( \vv{m} \rightarrow \vv{m}') } \\
&= \exp \left[ -\frac{1}{2} \left( \transpose{\vv{\delta}'} (b(\vv{m}')\mm{\Sigma})^{-1}  \vv{\delta}' - \transpose{\vv{\delta}} (b(\vv{m})\mm{\Sigma})^{-1}  \vv{\delta} - T\log\left(\frac{b(\vv{m})}{b(\vv{m}')}\right)  \right) \right] \cdot \frac{Z(\vv{m})}{Z(\vv{m}')} \\
&\approx \exp \left[ -\frac{1}{2} \left( \transpose{\vv{\delta}'} (b(\vv{m}')\mm{\Sigma})^{-1}  \vv{\delta}' - \transpose{\vv{\delta}} (b(\vv{m})\mm{\Sigma})^{-1}  \vv{\delta} - T\log\left(\frac{b(\vv{m})}{b(\vv{m}')}\right) \right) \right] ,
\end{align*}
where 
$\vv{\delta}' = \vv{s}-\mm{A}_0 \vv{m}'$, 
$\vv{\delta} = \vv{s}-\mm{A}_0 \vv{m}$, and the approximation is valid when the jump distance $\| \vv{m} - \vv{m}'\|$ is small enough for $Z(\vv{m}) \approx Z(\vv{m}')$.
If this is not the case or the composition vector is near the boundary of the simplex, then the approximation may not be good enough. Then we need to evaluate the constants empirically using MCMC methods. Another strategy to ensure the convergence of the Markov chain is to  selectively evaluate the constants if the distance $\| \vv{m} - \vv{m}'\|$ is larger than a pre-defined threshold. 
The expected benefit of using this sampling strategy is the high acceptance rate when the training uncertainty is small, thus efficient sampling is performed. In other words, the proposal distribution would look similar to the target distribution because $b$ is close to one when $m_j^2 \ll 1$ and $r$ is small. 

\section{Experiments}
\label{ssec:synthetic}

\begin{figure}
  \centering
\begin{tabular}{ccc}
\includegraphics[width=4.5in]{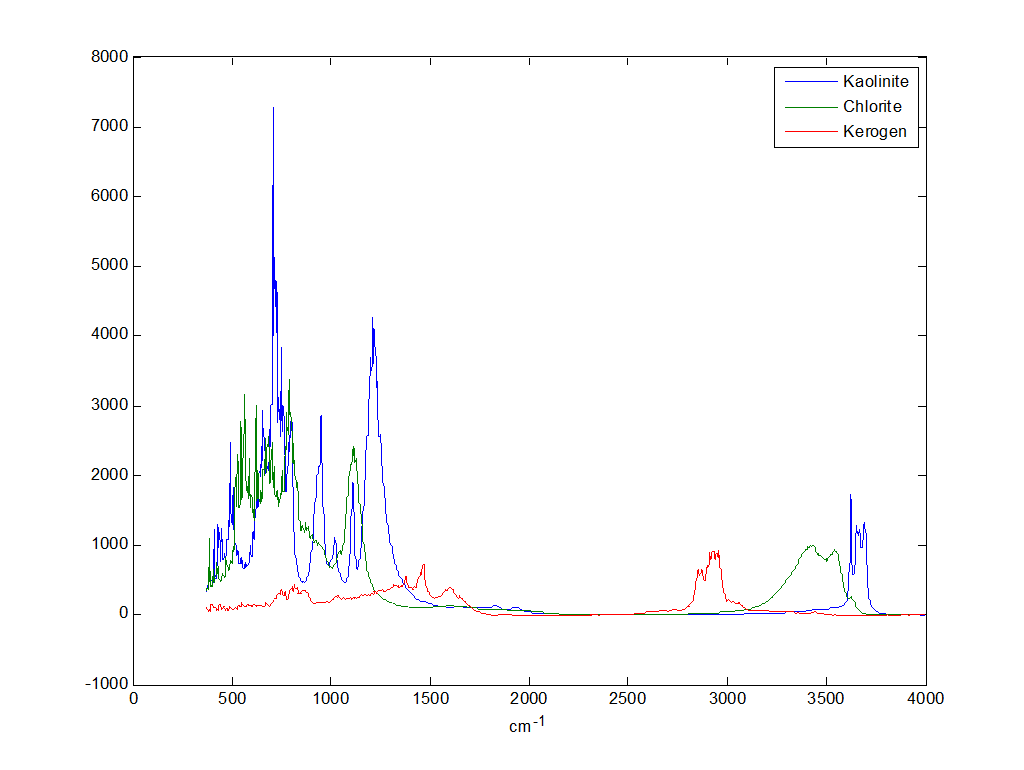} 
\end{tabular}
\caption{\label{fig:3easyMins}
Three easily separable spectra where three peaks above  2500 cm$^{-1}$ are distinguishable. The covariance (not shown) we use in our examples is a diagonal matrix and the noise level is high below  1500 cm$^{-1}$, thus suppressing signals in the region.
}
\end{figure}

\begin{figure}
  \centering
\begin{tabular}{ccc}
\includegraphics[width=4.5in]{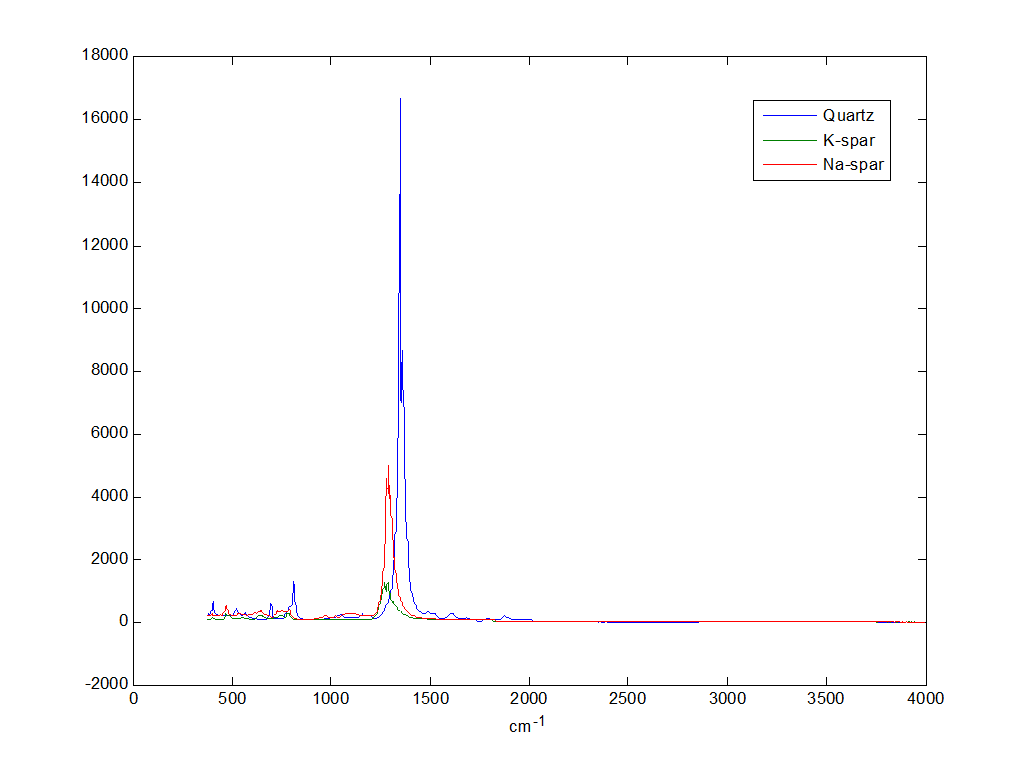} 
\end{tabular}
\caption{\label{fig:3noteasyMins}
Three mineral spectra that are difficult to separate. All the peaks are under the highly noisy region before 1500 cm$^{-1}$ and alike in terms of their locations. 
}
\end{figure}

We performed several tests of the proposed training-based inversion method on synthetically generated data. We consider a linear spectral system for our simulations. For testing and visualization, we use three endmembers ($M=3$) for training and inversion, and $T=2560$ (realistic dimension for high resolution spectral system) and used outputs from pure endmembers.
One dataset is built by using %
 synthetic spectra of pure minerals which have large weighted-$\ell_2$ spectral distances, so these minerals are expected to be easily separable. The weighted-$\ell_2$ distance between mineral spectra \vv{a} and \vv{b} is defined as $d_{\mm{\Sigma}^{-1}}(\vv{a},\vv{b}) = \transpose{(\vv{a} - \vv{b})} \mm{\Sigma}^{-1} (\vv{a} - \vv{b})$ where $\mm{\Sigma} $ is the noise covariance. The noise covariance $\mm{\Sigma}$ can be estimated from the training set, after generating noises using the true noise covariance. %
Fig.~\ref{fig:3easyMins} presents the spectra of these easily separable kaolinite, chlorite, and kerogen because their three peaks after 2500 cm$^{-1}$ do not overlap. 
The other dataset is fabricated with %
the spectra of minerals which have small weighted-$\ell_2$ distances, so they are difficult to separate. Fig.~\ref{fig:3noteasyMins} presents the spectra of minerals (quartz, K-spar, and Na-spar) that look difficult to separate. The spectra overlap in the wavenumber range of peaks and look alike in their shape.

\subsection{Minerals that are easy to separate, with $\vv{m} = [1/3, \,\,  1/3, \,\, 1/3]$}

\begin{figure}
  \centering
\begin{tabular}{ccc}
\includegraphics[width=4.5in]{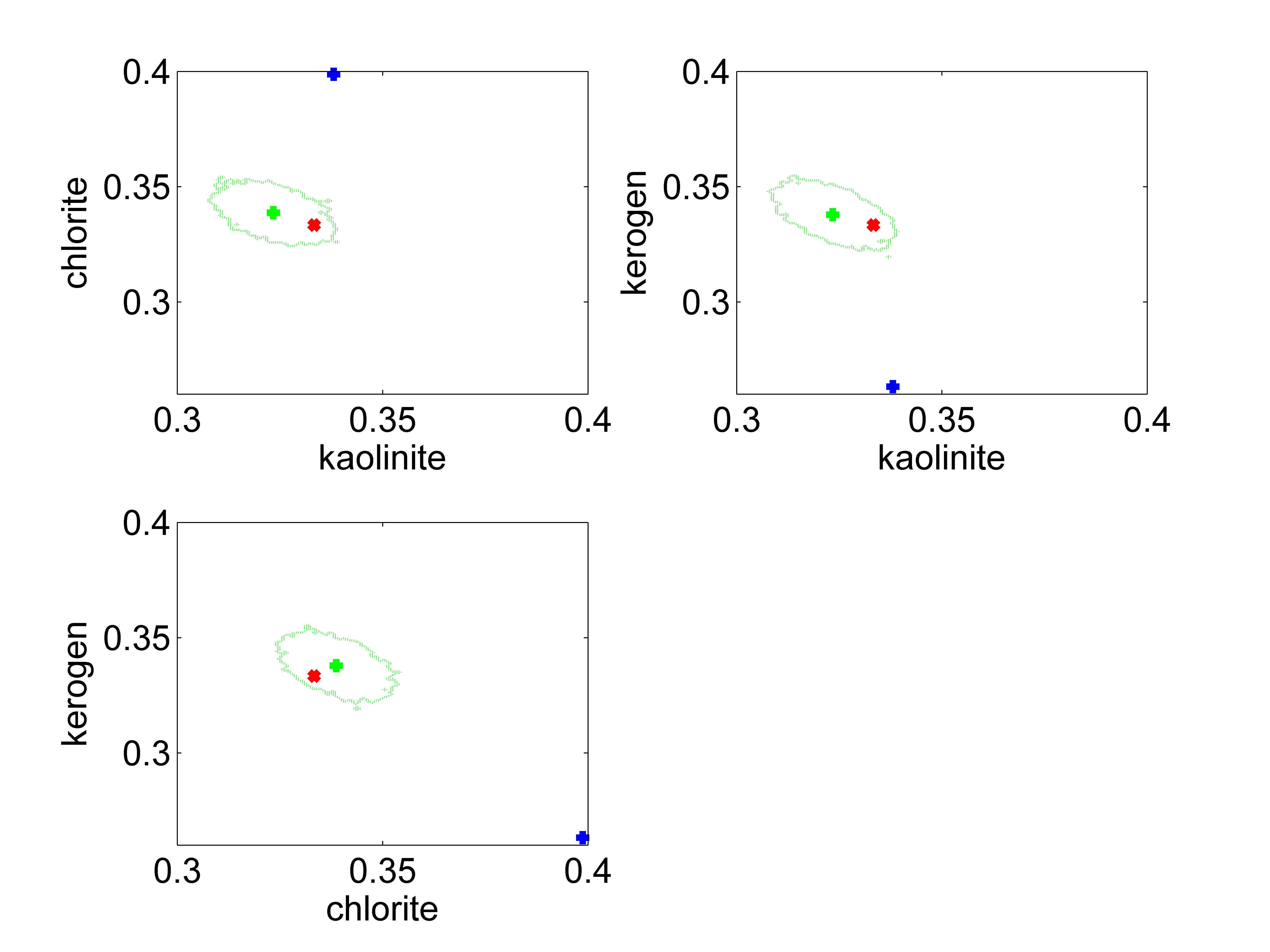} 
\end{tabular}
\caption{\label{fig:syntheticExN30_easySeparable}
The inversion of easily separable spectra from Fig.~\ref{fig:3easyMins} using a training set of $N=30$. The MAP estimate is the green dot, 95\% confidence interval is the green ellipse, true mineralogy is the red dot, and the blue dot is the estimated mineralogy without considering the uncertainty coming from estimation of \mm{A}. This result demonstrates that the uncertainty coming from the training of \mm{A} must be considered to compute the solution and avoid the large bias. Our CI (green ellipse) contains the true solution (red dot). We note that the uncertainty level is approximately $\pm 0.01$ or 1 weight percent.
}
\end{figure}

The first test data has a training set of $N=30$ easily separable end-members (pure minerals) from Fig.~\ref{fig:3easyMins}, and the true mineralogy is $\vv{m} = [1/3, \,\,  1/3, \,\, 1/3]$. The corresponding inversion result
 addressing the uncertainty in the trained $\mm{A}$ is presented in Fig.~\ref{fig:syntheticExN30_easySeparable}. %
We present the 2-dimensional sample 95\% confidence interval (CI) or ellipse in green curve and our CI contains the true solution, whereas the solution that does not consider the uncertainty coming from the training of \mm{A} has a large bias. 
Also, with various values of $N$ we empirically observed that the confidence ellipse (from the sample covariance structure) converges to the model, true covariance as the size of the training set $N$ increases.

One significant advantage of our proposed method with the stochastic \mm{A} model is visualized in Fig.~\ref{fig:syntheticExN30_easySeparable} by comparing our solution (MAP and CI in green dot and ellipse, respectively) with the result  from the conventional optimization analysis in blue dot. By `conventional' approach, we mean the optimization approach with \mm{A} fixed, as described in Section \ref{sssec:solAfixed}. In this conventional approach, $\mm{A}$ is the MLE and treated as a fixed parameter playing a similar role of  $\mm{A}_0$ in the stochastic \mm{A} model. 
This inversion is performed by maximizing the posterior density (\ref{eq:post2}), which is equivalent to maximizing (\ref{eq:likelihood}) with the same constraint, $\sum m_i = 1, m_i \geq 0$. 
Since the estimation problem with fixed \mm{A} can be posed as a convex programming, we used a software package, Matlab fmincon, with the constraint %
in maximizing the likelihood function, which produces a point estimate.
The inversion result %
clearly indicates the need for our approach. Moreover, %
because of the large bias, the probable solution area from the fixed \mm{A} does not cover the true mineralogy unlike ours\footnote{This is not shown to avoid convoluted graphics but it is trivial to see that the solution is far away from the true mineralogy}. %
 The comparison concludes that without addressing the uncertainty of the standards $\mm{A}$ obtained from the training, we may not be able to obtain solutions whose  confidence interval contains the true mineralogy.

\subsection{Minerals that are difficult to separate, with $\vv{m} = [1/3, \,\,  1/3, \,\, 1/3]$}

A similar test set is constructed (end-members in the training set, $ \vv{m} = [1/3, \,\,  1/3, \,\, 1/3]$) but with minerals of quartz, K-spar, and Na-spar that are difficult to separate (Fig.~\ref{fig:3noteasyMins}). Here, we used $N=1000$ training samples due to the low resolution of the spectra in Fig.~\ref{fig:3noteasyMins}. The low resolution is caused by the overlapping peaks of the quartz, K-spar, and Na-spar, as well as the their peak locations in noise abundant region ($< 1500$ cm$^{-1}$). We also note that for $N=1000$ the sample noise covariance is close to the true covariance, thus verifying the theory of the estimation of the covariance matrix in Section \ref{ssec:noiseCov}.

\begin{figure}
  \centering
\begin{tabular}{ccc}
\includegraphics[width=4.8in]{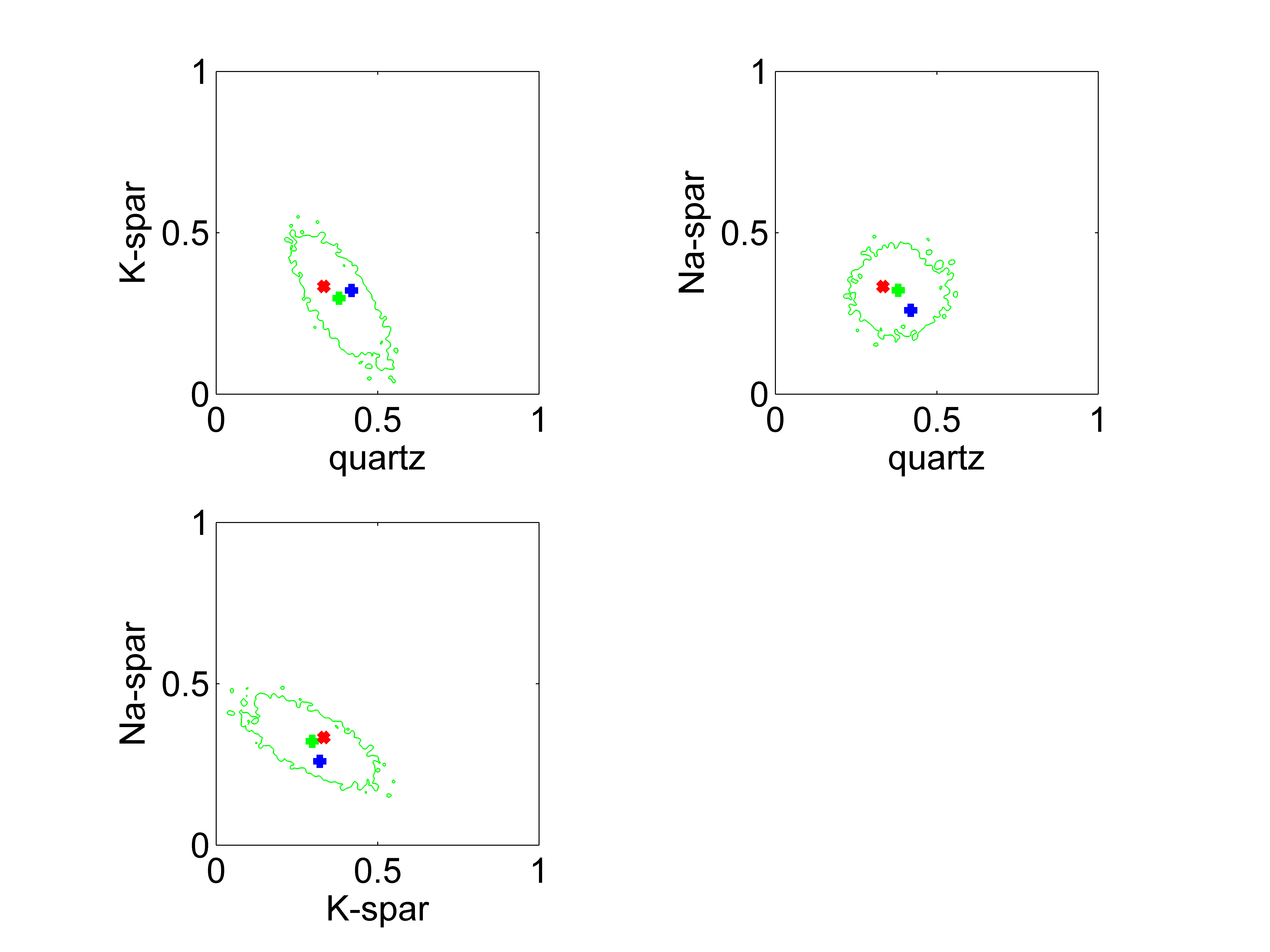} 
\end{tabular}
\caption{\label{fig:syntheticExN1000_NoneasySeparable}
Inversion result using $N=1000$ training samples and the three spectra that are difficult to separate from Fig.~\ref{fig:3noteasyMins}. 
The MAP estimate is the green dot, 95\% confidence interval is the green ellipse, true mineralogy is the red dot, and the blue dot is the estimated mineralogy without considering the uncertainty coming from estimation of \mm{A}. 
 Again, our CI (green ellipse) contains the true solution (red dot). The solution not considering the uncertainty from training is close to the true mineralogy. However, the uncertainty level now is approximately $\pm 0.1$ or 10 weight percent, almost 10 times larger than the case of easily separable minerals, compared to Fig.~\ref{fig:syntheticExN30_easySeparable}. The separability of the minerals caused this large uncertainty. 
}
\end{figure}

In the inversion result presented in Fig.~\ref{fig:syntheticExN1000_NoneasySeparable}, 
our CI (green ellipse) again contains the true solution (red dot). This time the solution not considering the uncertainty from training is close to the true mineralogy. However, the uncertainty level now is approximately $\pm 0.1$ or 10 weight percent, almost 10 times larger than the case of easily separable minerals, compared to Fig.~\ref{fig:syntheticExN30_easySeparable}. The separability of the minerals caused this large uncertainty.
The size of the CI we evaluated can be thought of the uncertainty level of mineralogy; if the noise level is small, the size of the confidence ellipse derived from the mineralogy covariance shrinks, and if the minerals are easily separable, e.g., major non-overlapping peaks over 2000 cm$^{-1}$, the covariance and ellipse size shrinks.

\subsection{Minerals that are difficult to separate, with $\vv{m}$ from a uniform distribution on $S_3$}

\begin{figure}
  \centering
\begin{tabular}{ccc}
\includegraphics[width=3.8in]{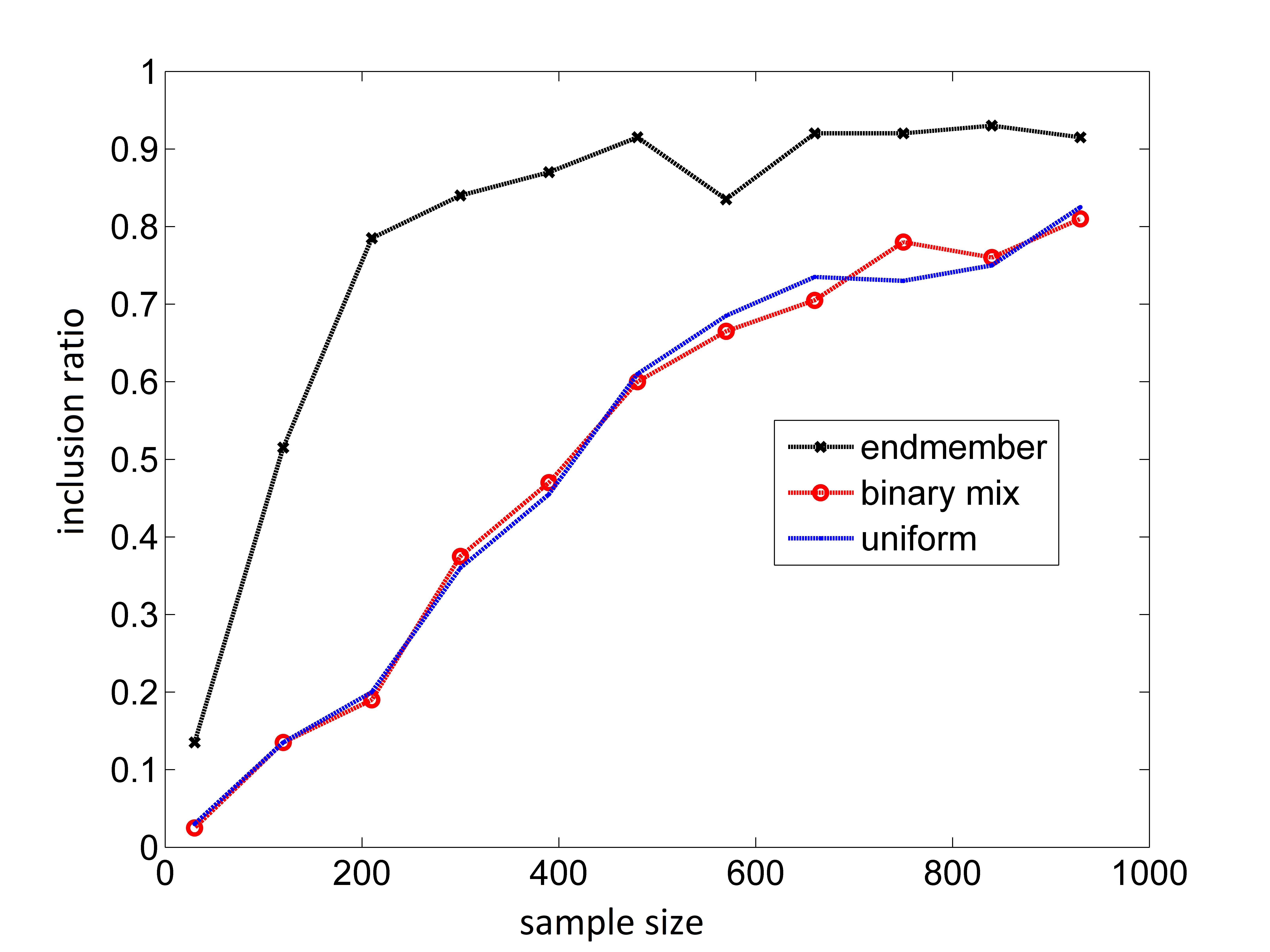} 
\end{tabular}
\caption{\label{fig:InclusionRateOverN_several}
The inclusion ratio of the true mineralogy within 95\% CI for all minerals, considering the three spectra that are most difficult to separate  (Fig.~\ref{fig:syntheticExN30_easySeparable}). In the simulation, the true mineralogy is uniformly drawn on $S_3$. The performances presented here verifies our prediction based on the error factor $r$. For pure mineral training samples (\vv{m} = end-member), $r = M/N = 3/N$, for binary mix training samples, $r = (2M + \frac{M}{M-1})/N = 9/N$, and for uniformly drawn training samples, $r=M^2/N = 9/N$. Therefore, the pure mineral set leads to the least error factor and the performance for the remaining two cases is comparable. 
}
\end{figure}

To link the training error to the performance of the inversion in the worst case, we synthesize data with the following setting: we uniformly draw a sample as a true mineralogy made of three minerals difficult to separate (quartz, K-spar, Na-spar) from $S_3$, later to test whether it is within 95\% CI from our inversion. Under the uniform distribution, all the valid mineralogy vectors have the same chance of selection/sampling. 
For the training set, we obtain 
$N$ samples under the several structures: pure minerals, binary mixes, and uniformly distributed mineralogy. Under each structure we perform the inversion and evaluate the `inclusion ratio' of the true mineralogy within 95\% CI for all minerals.  We vary $N$ from 30 to approximately 1,000.  %
In Fig.~\ref{fig:InclusionRateOverN_several},
the evaluated performances as of inclusion ratio verify our prediction based on the error scaling factor $r$. For pure mineral training samples (\vv{m} = end-member), $r = M/N = 3/N$, for binary mix training samples, $r = (2M + \frac{M}{M-2})/N = 9/N$, and for uniformly drawn training samples, $r=M^2/N = 9/N$. Therefore, the pure mineral case performs the best and the other two cases perform similarly and this was forecasted before inversion using $r$ values.

\subsection{Discussions}

Our proposed method can address uncertainty in mineralogy space, which comes from the three sources of the spectral uncertainty -- spectral noise, the ambiguity/similarity of certain minerals such as feldspar and quartz, and the training error. The noise directly affects the degree of uncertainty in the inversion whereas the ambiguity of minerals requires a study in their individual spectra. The training error can be predicted and it will be resolved by using a large number of the training samples. To resolve this error, from our experience, the minimum $N$ should lead to $r<0.01$ under a certain structure of the training set, thus an $N$ value can be suggested. 
We performed a fundamental analysis of ambiguity of minerals by using the weighted distance measure on the standards to understand which minerals are difficult or easy to separate and their effect in reconstructed mineralogy.
From our numerous experiments, we found that when the estimation of the mineral standards \mm{A} is close to the true mineral standards $\mm{A}_0$, e.g. $r<0.015$ in the worst test case, the inversion result is reliable even when the conventional approach with the fixed \mm{A} model is applied. Otherwise, it is better to apply the stochastic \mm{A}  model for the unbiased estimator. The estimation of $\mm{A}_0$ indeed affects the performance of uncertainty analysis in inversion. We can verify this effect by checking whether the 95\% confidence interval centered at the estimated mineralogy captures the true mineralogy and by looking at the estimated covariance structure converging to the true covariance structure as the size of the training set increases.

\section{Conclusion} 
\label{sec:conclusions}

We propose a novel training and inversion approaches for compositional variables.
The developed theory addresses several aspects of uncertainty that can arise in analyzing linear forward  models, especially regarding training of the system and reconstruction of the unkown compositions.
The proposed training analysis quantifies the variance scaling factor for the estimation of \mm{A}. We evaluated this factor under several cases of either deterministic or stochastic training sets, thus accounting for the uncertainty propagated from the estimated standards in the training step. 
The derived posterior distribution for a compositional variable produce more robust solutions that can capture uncertainty or ambiguity of compositional vectors. 
Especially, the comparisons between using fixed and stochastic \mm{A} models, when there is an uncertainty in \mm{A}, demonstrate that 
without addressing the uncertainty of the estimated $\mm{A}$ obtained from the training, we may not be able to obtain a solution whose confidence interval contains the true composition. When the estimate \mm{A} %
is close to the true \mm{A} %
(for large $N$ or small $r$), then a point estimate obtained from fixed \mm{A} model can be accurate. Our theory implies that when $r$ is small, the uncertainty amplification factor in the estimated operator is close to unity ($b \approx 1$) and both the fixed and stochastic \mm{A} models behave identically.

\appendices

\section{Maximum bound for $b$ in the posterior marginal distribution $p(\vv{m} | \vv{s})$ }
\label{appen:bMax}

The maximum of $b$ can be found as follows
\begin{align}
b &:= 1 + \sum_{j=1}^M m_j^2 r_j \\
 &\leq 1 + \sum_{j=1}^M m_j r_j  \\
 &\leq 1 + \max(r_j) =: b_{max} .
\end{align}
The first inequality holds because $m_j^2 \leq m_j $ for $m_j \in [0, 1]$. The second inequality holds because $\sum_{j=1}^M m_j r_j \leq \max(r_j ) \sum_{j=1}^M m_j $ and $\sum_{j=1}^M m_j = 1$. 

Furthermore, $b_{max}$ is the tight bound for $b$. For example, if the sample is a pure endmember ($m_k = 1$ for some $k$) and the corresponding variance scaling factor for the endmember is the largest ($r_{max} = r_k$), then  $b = 1 + \sum_{j=1}^M m_j^2 r_j = 1+ m_k^2 r_k = 1 + r_k = b_{max}$. Therefore, the maximum is achieved in this case.

\section{Distributions on an $n$-simplex}
\label{appen:distSimplex}

The composition vector $\vv{m}$ is constrained such that its components are nonnegative and sum to unity. 
These constraints define a simplex set such that any feasible \vv{m} is in the simplex set. 
An $n$-dimensional simplex, or $n$-simplex, is defined by 
\begin{equation}
\label{eq:def_simplex}
S_n = \{ (x_1, \ldots , x_n) \in \bb{R}^n \, : \,  \sum_{i=1}^n x_i = 1, x_i \geq 0 \mbox{ for } \forall i \} .
\end{equation}
Any feasible $\vv{m}$ of length $M$ is in $S_M$. 
Let the random members of \vv{m} be denoted $\vv{m} = \transpose{[U_1, U_2, \ldots, U_n]}$, where the $U_i$ are random variables. 

\subsection{Correlation property of distributions on an $n$-simplex } 
\label{ssec:PropertiesDistOnSimplex}

For $\vv{m} = \transpose{[U_1, U_2, \ldots, U_n]} \in S_n$, we define the following: 
\begin{align}
\label{eq:defYn}
Y_k &= \sum_{i=1}^k U_i ,\\
\label{eq:defYn2}
d_k &= \bb{E} [Y_k^2], \mbox { and } \\
\label{eq:defYn3}
\sigma &= \bb{E} [U_1^2].
\end{align} 
$Y_k, d_k, \sigma$ are dependent on $n$, however for simplicity we omit this dependence in our notation.
From the definition of the $n$-simplex, it is clear that 
\begin{align}
\label{eq:Yn1}
Y_n &= 1, \\
\label{eq:Yn2}
d_n &= 1, \mbox{ and } \\
\label{eq:Yn3}
\sigma &= d_1.
\end{align} 
These can be considered as boundary conditions for a distribution on an $n$-simplex.

One can derive a recursive equation for $d_k$ under the assumptions (\ref{eq:E_assumption3}):
\begin{align}
\label{eq:dk_recur}
d_k &= \E[U_1^2] + \E[Y_{k-1}^2] + 2\E[U_1 ( U_2 +  \cdots + U_k)] \\
&= \E[U_1^2] + d_{k-1} + 2(k-1) \E[U_1  U_2] \\
&= \sigma + d_{k-1} + 2(k-1) \beta.
\end{align}
Applying this recursion to solve for $d_k$ yields %
\begin{align}
\label{eq:dk_closed}
d_k = k \sigma  +  (k-1)k \beta .
\end{align}
Applying (\ref{eq:Yn2}) into (\ref{eq:dk_closed}) when $k=n$ leads to 
\begin{align}
\label{eq:dk_boundary_cond04}
1 = n \sigma  +  (n-1)n \beta
\end{align}

This equation (\ref{eq:dk_boundary_cond04}) relating $\sigma$ and $\beta$ can be used to demonstrate an intuitive and interesting property on the correlation. 
The correlation between random variables $U_1$ and $U_2$ is defined as follows:
\begin{align}
\label{eq:defCorr}
Corr(U_1,U_2) = \frac{\E\left[ (U_1 - \E U_1)(U_2 - \E U_2) \right]}{\sqrt{\E(U_1 - \E U_1)^2}\sqrt{\E(U_2 - \E U_2)^2}}
\end{align}
Under the following assumption, %
\begin{equation}
\label{eq:E_assumption2}
\bb{E} [U_j] = \bb{E} [U_k] =\mu \mbox{ for } j \neq k,
\end{equation}
the mean is 
\begin{align}
\label{eq:muU1}
\mu = 1/n .
\end{align} 
Using (\ref{eq:dk_boundary_cond04}) and (\ref{eq:muU1}), (\ref{eq:defCorr}) becomes
\begin{align}
\label{eq:Corr2}
Corr(U_1,U_2) &= \frac{\beta - \mu^2}{ \sigma - \mu^2 } \\
&= - \frac{1}{n-1} 
\end{align}
This result matches our intuition;  when a variable is increased by $\Delta$, the sum of the $(n-1)$ remaining variables is decreased by $\Delta$, thus the expectation of the decrease in one of the remaining variables would be $\Delta/(n-1)$.

\subsection{Example distributions on an $n$-simplex}
\label{ssec:distOnSimplex}

We consider three distributions on a $n$-simplex that satisfy  (\ref{eq:E_assumption3}) and (\ref{eq:E_assumption2}): the multinomial distribution $Mult(n)$, the double-multinomial distribution $Mult_2(n)$, and the uniform distribution $ Unif(n)$. 

\subsubsection{Multinomial distribution }
\label{sssec:mult}

If $\vv{m} \in \bb{R}^M \sim Mult(M)$, then 
\begin{equation}
\label{eq:defMult}
P(\vv{m}=\vv{e}_j) = 1/M \mbox{ \space for } j = 1,\ldots,M,
\end{equation}
where $\vv{e}_i$ is the $i$-th column of the identity matrix.
Under this distribution,  
\begin{equation}
\E[U_1 U_2] = 0 \mbox{ and } \E U_1^2 = \E U_1 = 1/M, 
\end{equation}
leading to 
\begin{equation}
\label{eq:r_Mult}
r = M/N . 
\end{equation}
This result is comparable to the repeated-mixtures case given in Eq.~(\ref{eq:r_rep}) with $\mm{M}_0 = \mm{I}_M$ and $K  = N/M$.

\subsubsection{Double-multinomial distribution with or without replacement }
\label{sssec:dmult}

This distribution is similar to the case of multinomial distribution, but endmember selection is performed twice. 
If endmembers are selected with replacement, the same endmember may be selected twice for a given sample, resulting in the corresponding column of \mm{M} having a single non-zero entry of unity, or if distinct endmembers are selected, there will be two non-zero entries of $1/2$. If entries are selected without replacement, two distinct endmembers are always selected, and the columns of \mm{M} will always have two non-zero entries of $1/2$.
Thus, we define the probability mass functions as follows:

Double-multinomial distribution with replacement: 
\begin{equation}
\label{eq:def_dmultW}
P (\vv{m} = \frac{1}{2} (\vv{e}_k + \vv{e}_l) ) = \frac{1}{M^2} \left(2\bb{I}(k \neq l)+\bb{I}(k = l) \right) \, \mbox{ for } k,l \in [1,\ldots, M] . 
\end{equation}

Double-multinomial distribution without replacement:
\begin{equation}
\label{eq:def_dmultWo}
P (\vv{m} = \frac{1}{2} (\vv{e}_k + \vv{e}_l) ) = 2/(M^2-M) \, \mbox{ for } k \neq l \in [1,\ldots, M] . 
\end{equation}

For the case with replacement,
\begin{align}
\label{eq:der_dmultW}
\E[U_1^2] &= (1)^2P( U_1 = 1 ) + (\frac{1}{2})^2 P( U_1 = \frac{1}{2}) \nonumber\\
 &= \frac{1}{M^2} + \frac{1}{2^2} \frac{2(M-1)}{M^2}  \nonumber\\
 &= \frac{M+1}{2M^2}
\end{align}
 and
 \begin{align}
 \label{eq:der_dmultW2}
\E[U_1 U_2] &= (\frac{1}{2})^2  P( U_1 =  \frac{1}{2}, U_2= \frac{1}{2}) \nonumber\\
 &= \frac{2}{4M^2},
\end{align}
leading to 
\begin{align}
\label{eq:r_der_dmultWo}
r=\frac{2M-1}{N} .
\end{align}

For the case without replacement,
\begin{align}
\label{eq:der_dmultWo}
\E[U_1^2] &=  (\frac{1}{2})^2 P( U_1 = \frac{1}{2}) \nonumber\\
 &= \frac{1}{2^2} \frac{2(M-1)}{M(M-1)} \nonumber\\
 &= \frac{1}{2M}
 \end{align}
 and
 \begin{align}
\label{eq:der_dmultWo2}
\E[U_1 U_2] &= (1/2)^2  P( U_1 = 1/2, U_2=1/2) \nonumber\\
 &= \frac{1}{2M(M-1)},
\end{align}
leading to 
\begin{align}
\label{eq:r_der_dmultWo}
r=\frac{1}{N}(2M + \frac{M}{M-2} ) .
\end{align}

Comparing the above scaling factors (linear with $M$) with that from repetitive binary mixtures (quadratic with $M$), we conclude that stochastically generated binary mixtures offer significantly improved variance.

\subsubsection{Uniform distribution}
\label{sssec:unif}

If $\vv{m} \in \bb{R}^n \sim Unif(S_n)$, then for any $\set{X} \subset S_n$ 
\begin{equation}
\label{eq:defUnifSn}
P(\vv{m} \in \set{X} ) = \frac{Vol(\set{X})}{Vol(S_n)},
\end{equation}
where $Vol(\set{X})$ is a Lebesque integral on $\set{X}$. 
Therefore, $P(\vv{m} \in \set{X} )$ is independent of a `location' of a set $\set{X}$ and is a uniform distribution on $S_n$. Call $p(\vv{u})$ a probability density function (pdf) for a random vector $\vv{U}$. 

For example, when $n=2$, it is easy to show 
\begin{align} 
\E[U_1^2] = 1/3 \\
\E[U_1 U_2] = 1/6
\end{align}
When $n=3$ ($\vv{u} \in \bb{R}^3$), note that 
\begin{align} 
p(\vv{u}) =  \frac{\bb{I}(\vv{u} \in S_3)}{Vol(S_n)} , 
\end{align}
\begin{align} 
p_{3 \rightarrow 2} (u_1, u_2) & :=  \int p(\vv{u}) du_3 \\
&= \bb{I} ((u_1,u_2) \in V_2) / Vol(V_2), 
\end{align}
where one can show $Vol(V_n) = 1/n!$
and 
\begin{align} 
p_{3 \rightarrow 1} (u_1) & :=  \int p(\vv{u}) du_2 du_3 \\
&= (1-u_1) \bb{I} (u_1 \in [0,1]) / Vol(V_2) \\
&= 2(1-u_1) \bb{I} (u_1 \in [0,1]) .
\end{align}
Using the above findings, 
\begin{align} 
\E[U_1^2] &= \int_{S_3} u_1^2 p(\vv{u}) \,\, du_3 du_2 du_1 \\
 &= \int_{[0,1]} u_1^2 \, p_{3 \rightarrow 1} (u_1) du_1 \\
 &= \int_{[0,1]} u_1^2 \cdot  2(1-u_1) du_1 \\
 &= 1/6 
\end{align}
Similarly,
\begin{align} 
\E[U_1 U_2] &= \int_{S_3} u_1 u_2 \, p(\vv{u}) \,\, du_3 du_2 du_1 \\
 &= \int_{V_2} u_1 u_2 \, p_{3 \rightarrow 2} (u_1,u_2) du_2 du_1 \\
 &= \int_{[0,1]} \left( \int_{[0,1-u_1]} u_1 u_2 \, \frac{1}{Vol(V_2)} du_2 \right) du_1 \\
 &= 1/12
\end{align}


Generally, one can show the followings:
\begin{align} 
p_{n \rightarrow n-1}(\vv{u}^{n-1}) &= \frac{1}{Vol(V_{n-1})} \bb{I}(\vv{u}^{n-1} \in V_{n-1})  \\
p_{n \rightarrow n-2}(\vv{u}^{n-2}) &= \frac{1}{Vol(V_{n-1})} \bb{I}(\vv{u}^{n-2} \in V_{n-2}) (1 - \sum_{i=1}^{n-2}u_i) \\
p_{n \rightarrow n-3}(\vv{u}^{n-3}) &= \frac{1}{Vol(V_{n-1})} \bb{I}(\vv{u}^{n-3} \in V_{n-3}) \dfrac{(1 - \sum_{i=1}^{n-3}u_i)^2 }{2} \\
p_{n \rightarrow n-4}(\vv{u}^{n-4}) &= \frac{1}{Vol(V_{n-1})} \bb{I}(\vv{u}^{n-4} \in V_{n-4}) \dfrac{(1 - \sum_{i=1}^{n-4}u_i)^3 }{2\cdot 3}  \\
\vdots \\
p_{n \rightarrow k}(\vv{u}^{k}) &= \frac{1}{Vol(V_{n-1})} \bb{I}(\vv{u}^{k} \in V_{k}) \dfrac{(1 - \sum_{i=1}^{k}u_i)^{(n-1-k)} }{ (n-1-k)!} , 
\end{align}
where $\vv{u}^{s} := (u_1, ... , u_{s})$.
Therefore, the terms of interest are:
\begin{align} 
p_{n \rightarrow 2}(\vv{u}^{2}) &= \frac{1}{Vol(V_{n-1})} \bb{I}(\vv{u}^{2} \in V_{2}) \dfrac{(1 - u_1 - u_2 )^{(n-3)} }{ (n-3)!}  \\
p_{n \rightarrow 1}(\vv{u}^{1}) &= \frac{1}{Vol(V_{n-1})} \bb{I}(\vv{u}^{1} \in V_{1}) \dfrac{(1 - u_1)^{(n-2)} }{ (n-2)!}  \\
\end{align}
Using $(a+1)^n = \sum_{i=0}^n a^i \dfrac{n!}{i! (n-i)!}$ and the definition of expectation, the followings hold:
\begin{align} 
\label{eq:solved_exp}
\E [U_1^2] = \frac{2}{n(n+1)} \\
\E [U_1 U_2] = \frac{1}{n(n+1)} \\
\end{align}
Therefore, $\alpha = \beta = \dfrac{1}{n(n+1)} $. By substituting $n=M$ and $\alpha,\beta$ into \eqref{eq:vrf} the variance reduction factor is 
\begin{align} 
\label{eq:vrf_uniform}
r = \frac{M^2}{N}
\end{align}

\subsubsection{Pseudo-uniform distribution}
\label{sssec:punif}
The use of this distribution is motivated by its easy nature of generating samples. 
If $\vv{m} \in \bb{R}^n \sim pUnif(n)$, then there exist random variables $\{ U'_i \}_{i=1}^n$ such that 
\begin{align}
\label{eq:defpUnif}
\vv{m} = \transpose{[U'_1, U'_2, ..., U'_n]} / \sum_{i=1}^n U'_i,
\end{align}
where each $U'_i$ is independently and identically distributed and follows a uniform distribution on $[0, 1]$. This distribution looks close to the uniform distribution but is slightly more centralized than the uniform distribution. However, we note that the sampling procedure on this distribution is easier, by using uniform random variables on [0,1], than on the uniform distribution over $n$-simplex. 

The moments used in evaluating $r$ can be numerically computed. 
Under this distribution, 
\begin{align}
\label{eq:pUnifProp}
\E[U_1 U_2] &= \int_{[0,1]^n} \frac{u_1 u_2}{(u_1 + u_2 + ... + u_n)^2} du_1 du_2 \cdots du_n \\
\E U_1^2 &= \int_{[0,1]^n} \frac{u_1^2}{(u_1 + u_2 + ... + u_n)^2} du_1 du_2 \cdots du_n . 
\end{align}


\bibliographystyle{IEEEtran}
\bibliography{DRIFTSxFTIR03}
\end{document}